\documentclass[11pt]{article}

\usepackage[preprint]{acl}

\usepackage{times}
\usepackage{latexsym}
\usepackage{amsmath}
\usepackage{booktabs}
\usepackage{minted}
\usepackage{subcaption}
\usepackage{listings}

\usepackage[T1]{fontenc}

\usepackage[utf8]{inputenc}

\usepackage{microtype}

\usepackage{inconsolata}

\usepackage{graphicx}

\title{Uncertainty-Aware Trust Estimation for Multi-LLM Systems via Structured Expert Judgement}

\author{Jiawei Zheng \and Jiazhen Zhang \\
        DigitLab, University of Exeter \\ \small{j.zheng2@exeter.ac.uk, j.zhang15@exeter.ac.uk} \\
}

\begin{document}
\maketitle
\begin{abstract}
Large Language Model (LLM) ensembles are increasingly used to improve reliability by combining predictions from multiple LLMs. However, existing aggregation methods typically assume that all models are equally trustworthy, overlooking differences in uncertainty quality. This assumption is poorly suited to heterogeneous LLMs, whose reliability and capability vary significantly, making naive aggregation vulnerable to unreliable or adversarial experts. In this work, we formulate multi-LLM aggregation as a problem of uncertainty-aware trust estimation. We adapt structured expert judgment from decision theory, using context-aware calibration questions to estimate expert reliability based on the quality of its probabilistic predictions. Specifically, we employ Cooke-style log weighting, which penalises overconfident incorrect predictions and favours well-calibrated experts. We evaluate our approach on MMLU and MMLU-Pro across homogeneous, heterogeneous, and contaminated expert panels. Results show that while aggregation methods perform similarly in homogeneous settings, Cooke weighting becomes critical under heterogeneity and contamination. It achieves a superior accuracy-reliability balance and remains robust when unreliable experts are introduced. These findings suggest that Multi-LLM aggregation requires not just combining predictions, but calibrating trust under uncertainty.
\end{abstract}

\section{Introduction}\label{sec:introduction}

Large language models (LLMs) are increasingly deployed in settings that require reliable reasoning under uncertainty, including scientific analysis, healthcare support, and legal decision-making~\cite{huangSurveyUncertaintyEstimation2024}. In these domains, the quality of a system depends not only on whether it produces the correct answer, but also on whether its confidence is justified~\cite{tianJustAskCalibration2023,liuUncertaintyEstimationQuantification2024}. Recent work has shown that modern LLMs frequently exhibit severe miscalibration, often producing confident yet incorrect predictions and hallucinations~\cite{kumaranCompetingBiasesUnderlie2026,lengTamingOverconfidenceLLMs2024,xiongCanLLMsExpress2023}. Such overconfident failures are particularly dangerous in high-stakes settings like healthcare, where users may rely on confidence signals to assess trustworthiness~\cite{zhouUncertaintyawareLargeLanguage2025,qinEnhancingHealthcareLLM2024}.

Accurately expressing uncertainty is essential for reliable and trustworthy employment of LLMs.
Prior work has studied uncertainty estimation and calibration in LLMs, showing that they are often miscalibrated and prone to overconfident errors~\cite{xiongCanLLMsExpress2023,tianJustAskCalibration2023,kumaranCompetingBiasesUnderlie2026}. 
However, these methods operate at the level of individual models, assuming that uncertainty can be addressed within a single model.

In practice, however, individual models exhibit varying reliability across different inputs and domains.~\cite{lincolnFewGoodClauses2026,kruse-etal-2025-simple,chenHarnessingMultipleLarge2025}. Different LLMs exhibit complementary strengths, generalising better in different regions of the input space due to differences in training data, objectives, and architectures~\cite{chanDataDistributionalProperties2022}. This diversity motivates combining multiple LLMs to improve robustness. However, it also raises a key challenge: models vary not only in accuracy, but in the quality of their uncertainty estimates~\cite{kumaranCompetingBiasesUnderlie2026}. 

Most existing aggregation methods, such as majority voting and equal averaging, remain accuracy-centric and treat all models as equally reliable, without accounting for differences in uncertainty quality~\cite{wangSelfConsistencyImprovesChain2022,chenHarnessingMultipleLarge2025}. As a result, poorly calibrated or overconfident models can disproportionately skew the final prediction.
One recent method estimates uncertainty from model disagreement~\cite{kruse-etal-2025-simple}, but does not directly address which models should be trusted. Consequently, unreliable or miscalibrated models may still exert substantial influence on the final prediction, particularly in heterogeneous or contaminated settings.

In this work, we frame multi-LLM aggregation as an uncertainty-aware trust estimation problem. Rather than asking how to combine predictions, we ask: which models should be trusted, and under what uncertainty conditions? 
To answer this, we adapt Cooke's method~\cite{cookeexperts1991,colsonExpertElicitationUsing2018}, a classical framework from decision theory for combining uncertain forecasts from different experts, to estimate the reliability of the LLM aggregation. 
Specifically, experts are evaluated using calibration (seed) questions with known answers, and weights are assigned using log-score rules that penalise overconfident mistakes while rewarding calibrated confidence.

We evaluate our approach on MMLU~\cite{hendryckstest2021} and MMLU-Pro~\cite{wangMMLUProMoreRobust2024} across three progressively challenging regimes: 1) homogeneous panels of frontier models, 2) heterogeneous panels combining models of varying capability, and 3) contaminated panels containing explicitly noisy or adversarial experts. In homogeneous settings, aggregation methods perform similarly. However, as heterogeneity and contamination increase, uncertainty-aware weighting becomes critical. In particular, Cooke weighting maintains strong accuracy while significantly reducing overconfident failures and preventing the severe performance degradation observed in other baselines.

\section{Related work}

\textbf{LLM ensembles} Combining multiple LLMs has emerged as an effective strategy to improve reliability and reasoning performance~\cite{choiDebateVoteWhich2025,aiMajorityVotingLLM2026}. In addition to the work mentioned in \S~\ref{sec:introduction}, a recent survey systematically categorises these methods into pre-, during-, and post-inference aggregation paradigms~\cite{chenHarnessingMultipleLarge2025}. While these approaches improve predictive performance, they typically assume equal model reliability or optimise towards accuracy~\cite{huangEnsembleLearningHeterogeneous2024}, without explicitly modelling differences in uncertainty quality across experts.

\noindent \textbf{Uncertainty quantification in LLMs} A growing body of work studies uncertainty estimation and calibration in LLMs, showing that LLMs are often miscalibrated and prone to overconfident errors~\cite{xiongCanLLMsExpress2023,tianJustAskCalibration2023,kumaranCompetingBiasesUnderlie2026}. However, most approaches focus on measuring or correcting uncertainty at the individual LLM level~\cite{riveraCombiningConfidenceElicitation2024,lengTamingOverconfidenceLLMs2024}. More recently, \cite{kruse-etal-2025-simple} proposes estimating uncertainty in multi-LLM systems using information-theoretic disagreement. They primarily quantify ensemble-level uncertainty from disagreement, rather than estimating the trustworthiness of individual LLMs.

\noindent \textbf{Structured expert judgement (SEJ)} Our work is closely related to SEJ, particularly Cooke’s method~\cite{cookeexperts1991}, which assigns weights to experts based on their performance on calibration questions using proper scoring rules.
We adapt SEJ to LLM by treating models as probabilistic experts and evaluating them based on the quality of their uncertainty estimates. Unlike prior LLM ensemble methods, which focus on combining predictions, our work focuses on uncertainty-aware trust estimation, enabling robust aggregation under heterogeneous and contaminated expert panels.
\section{Methods}

We treat a collection of LLMs as a panel of heterogeneous experts that produce probabilistic forecasts, and formulate aggregation as a problem of estimating trust under uncertainty.

\subsection{Problem Formulation}

We consider a panel of $M$ experts, where each expert $e_i$ produces probabilistic predictions for an input $x$. Let $\mathcal{Y}$ denote the output space, which may be discrete, structured, or sequence-based. For an input $x$, each expert provides a predictive distribution.
\begin{equation}
    p_i(y \mid x), \quad y \in \mathcal{Y}
\end{equation}

The goal is to aggregate these expert predictions into a single prediction while accounting for differences in expert reliability and uncertainty quality.

We assume access to a small set of calibration examples.
\begin{equation}
    \mathcal{D}_{seed} = \{(x_j, y_j)\}_{j=1}^N 
\end{equation}
where ground-truth outputs $y_j$ are known. These calibration examples are used to estimate the trust of each expert prior to aggregation on target sets.

\subsection{Probability elicitation}

For each input $x$, each expert produces a normalised predictive distribution. In practise, for LLMs, such distributions can be obtained through normalised probabilities over candidate outputs, or token-level likelihoods for sequence generation.

We assume that predictions satisfy:
\begin{equation}
    p_i(y \mid x) \geq 0, \quad \int_y p_i(y \mid x)\, dy = 1
\end{equation}

\subsection{Context-aware trust estimation}

We estimate expert reliability using the logarithmic scoring rule applied to calibration data. In many settings, expert reliability is not global but context-dependent. For example, an expert may perform well in one domain but poorly in another. To capture this, we introduce $c(x)$ representing a context variable associated with input $x$. The context variable is assumed to be observable and may be categorical (e.g., subject label) or derived from metadata or task structure.

For each expert $e_i$ and context $c$, we compute a context-specific score:
\begin{equation}
    s_{i,c}
=
\frac{1}{|\mathcal{D}_{seed}^c|}
\sum_{(x_i, y_j) \in \mathcal{D}_{seed}^c}
\log p_i(y_j \mid x_j),
\end{equation}
where $\mathcal{D}_{seed}^c = \{(x_j, y_j):c_j = c\}$ contains calibration examples in context $c$. When computing $log$ scores, the probabilities are clipped with a small constant $\epsilon$ to avoid numerical instability.

The logarithmic property ensures that overconfident incorrect predictions are penalised more heavily than uncertain ones. As a result, the scoring function can distinguish between calibrated uncertainty, where predictions reflect true uncertainty, and overconfident errors, where incorrect outcomes receive high probability. Consequently, trust estimation becomes sensitive to the quality of uncertainty, not just correctness.

The corresponding weights of the expert $e_i$ is computed by:
\begin{equation}
    w_{i,c} = \frac{
\exp(\tau S_{i,c})
}{
\sum_{m=1}^M \exp(\tau S_{m,c})
}
\end{equation}
where the temperature parameter \(\tau\) controls how sharply the method concentrates weight on high-scoring experts. Larger values of \(\tau\) increasingly favour the best-calibrated seed experts. We set $\tau$ as 1 unless explicitly mentioned. 

\subsection{Uncertainty-aware aggregation}

Given context-dependent weights of each expert, we compute the aggregated predictions as:
\begin{equation}
    p_{agg}(y \mid x)
=
\sum_{i=1}^M
w_{i,c}
p_i(y \mid x).
\end{equation}

The final prediction is derived from $p_{agg}$ depending on the task, e.g., argmax for classification or sampling for sequence generation.

\section{Experimental setup}

\subsection{Datasets}

We conduct experiments on MMLU and MMLU-Pro, two widely used benchmarks for assessing knowledge-intensive reasoning in large language models. MMLU consists of four-choice multiple-choice questions across a broad range of domains, including science, mathematics, academic, and professional subjects. MMLU-Pro is a more challenging extension of MMLU with more challenging questions and more subtle distinctions between answer choices. Each question contains up to ten answer choices. For each question, we instruct LLMs to output a probability distribution over choices, termed verbalised confidence~\cite{xiongCanLLMsExpress2023}. 

For each dataset, we evaluate a subset of subjects spanning the scientific, professional, mathematical, and social science domains. For each subject, we split questions into a \textit{seed set} and a \textit{target set}. Seed questions are used only to estimate expert reliability and compute aggregation weights. Target questions are used only for the final evaluation. Unless otherwise stated, we use a 20/80 seed-target split within each subject. We repeat experiments across 5 random splits to reduce variance.

\subsection{Expert panels}

We evaluate aggregation methods across three progressively challenging regimes designed to isolate the role of uncertainty-aware trust estimation.

\noindent \textbf{Homogeneous Frontier Panels}.
This setting contains only strong frontier proprietary models with relatively similar capability and calibration quality. The purpose of this regime is to evaluate whether weighting remains beneficial when expert quality is relatively uniform. The homogeneous panel consists of three frontier LLMs: 1) \textit{Claude Sonnet 4.6}, 2) \textit{Gemini 3 Flash Preview}, and 3) \textit{GPT-5.4}.

\noindent\textbf{Heterogeneous Expert Panels}
This setting combines frontier proprietary models and open-source models with varying capabilities and uncertainty characteristics. Unlike the homogeneous setting, expert quality varies substantially, creating a more realistic trust estimation problem.

Besides the three frontier models included in the homogeneous panel, this panel also contains: 1) \textit{Qwen 3.5 Plus}, 2) \textit{GPT-OSS-20b}, 3) \textit{Deepseek-v3.2}, and 4) \textit{Qwen 3.5 35b}. Therefore, this panel includes 7 experts in total. This configuration intentionally includes both strong and imperfect experts to evaluate how aggregation methods behave under reliability heterogeneity.

\noindent \textbf{Contaminated Expert Panels}
To further evaluate robustness under unreliable experts, we introduce contaminated expert panels containing synthetic noisy experts that simulate common failure modes in practical multi-LLM systems.

We consider three contamination types: 1) \textit{random expert}: returns a random probability distribution over choices, 2) \textit{overconfident wrong expert}: assigns high probability to an incorrect answer, and 3) \textit{biased expert}: systematically picks a fixed option regardless of the question.

To systematically study robustness under contamination, we vary the ratio of real LLM experts to noisy experts and introduce a contamination ratio, formally, $\rho = \frac{Y}{X + Y}$ where X is the number of real LLM experts and Y is the number of noisy experts. This setup enables controlled evaluation of how aggregation methods degrade as unreliable experts occupy a larger fraction of the panel.

\subsection{Baselines}

We compare the proposed Cooke trust estimation against solo models and several commonly used aggregation strategies, including:
\begin{itemize}
    \item Majority vote: Each expert votes for the highest-probability answer. The final answer is the most frequent vote. The vote count is converted into a probability distribution for probabilistic evaluation.
    \item Equal averaging: All experts receive equal weight, and their probability distributions are averaged.
    \item Global weighting: Each expert receives a single weight computed from its overall seed-set log score across all subjects. The same weight is used for every subject.
    \item Accuracy weighting: Each expert receives a subject-specific weight based on the seed-set accuracy for that subject. This baseline tests whether uncertainty-aware aggregation provides benefits beyond simple correctness.
\end{itemize}

\subsection{Evaluation metrics}

We evaluate each method along three dimensions: 1) accuracy, 2) probabilistic quality, and 3) risk under uncertainty. Accuracy measures whether the final answer with the highest probability is correct. Accuracy is the primary measure of answer correctness, but it discards information about the model's uncertainty. Two methods may achieve identical accuracy while assigning very different probabilities to the correct answer. We therefore also evaluate the predictive distribution assigned by experts. Risk-sensitive metrics further characterize whether a method is prone to confident mistakes.

We use \textit{negative log likelihood} (NLL) and multi-class \textit{Brier score} to evaluate the uncertainty quality~\cite{murphy2012machine}. 
To capture the risk of high-confidence failures, we also report \textit{overconfident error} (OE) rate.  We define an overconfident error as an incorrect prediction whose confidence exceeds a threshold $\gamma$. Formally:
\begin{equation}
\small
\text{OE}
=
\frac{1}{|T|}
\sum_{i \in T}
\mathbf{1}
[
\hat{y}_i \neq y_i^\ast
\land
\max_y p(y \mid x_i) > \gamma
]
\end{equation}
where $T$ means target sets, and we set \(\gamma = 0.7\) in all experiments. This metric captures a failure mode that may not be visible from accuracy alone: a method can be accurate on average but still produce a non-trivial number of high-confidence errors.

These metrics capture complementary aspects of uncertainty quality, including sharpness (NLL), calibration (Brier), and risk of overconfident errors (OE).

\section{Results}

\subsection{Homogeneous frontier panels}

\begin{table*}[htbp]
\centering
\small
\begin{tabular}{lcccccccc}
\toprule
& \multicolumn{4}{c}{MMLU} & \multicolumn{4}{c}{MMLU-Pro} \\
\cmidrule(lr){2-5} \cmidrule(lr){6-9}
Method & Acc $\uparrow$ & NLL $\downarrow$ & Brier $\downarrow$ & OE $\downarrow$
& Acc $\uparrow$ & NLL $\downarrow$ & Brier $\downarrow$ & OE $\downarrow$ \\
\midrule
Best solo model
& 93.75 & 0.304 & 0.119 & 5.77
& 87.77 & 0.605 & 0.233 & 3.34 \\

Majority vote
& \textbf{94.19} & \textbf{0.249} & \textbf{0.107} & 2.14
& 86.99 & 0.583 & 0.216 & 3.42 \\

Equal averaging
& 93.86 & 0.277 & 0.119 & \textbf{1.92}
& 86.02 & 0.516 & 0.225 & 2.42 \\

Global weighting
& 94.17 & 0.271 & 0.113 & \textbf{1.92}
& 86.86 & 0.502 & 0.212 & 2.35 \\

Accuracy weighting
& 94.14 & 0.273 & 0.116 & \textbf{1.92}
& 86.45 & 0.505 & 0.217 & 2.43 \\

\textbf{Cooke weighting}
& 94.04 & 0.265 & 0.109 & \textbf{1.92}
& \textbf{88.03} & \textbf{0.485} & \textbf{0.201} & \textbf{2.34} \\
\bottomrule
\end{tabular}
\caption{Results of homogeneous panels with three frontier models on MMLU and MMLU-Pro. Values are averaged over 5 runs; standard deviations are negligible and omitted (see Table~\ref{tab:apx:top3models} in Appendix). Best results are in bold.}
\label{tab:homogeneous-results}
\end{table*}

We first evaluate aggregation methods on homogeneous panels consisting exclusively of three strong frontier models. 

Table~\ref{tab:homogeneous-results} presents the results for the homogeneous frontier panel on MMLU and MMLU-Pro. The results show that across both benchmarks, all ensemble methods perform similarly in terms of accuracy. On MMLU, majority vote achieves the highest accuracy (94.19\%), with Cooke weighting close behind (94.04\%). On MMLU-Pro, Cooke weighting performs best overall, reaching 88.03\% accuracy and slightly outperforming the best solo model. This demonstrates that when expert quality is relatively homogeneous, aggregation provides limited gains over strong individual models.

Although accuracy differences are small, clearer distinctions emerge in probabilistic metrics. On MMLU-Pro, Cooke weighting achieves the best NLL and Brier score, indicating the strongest probabilistic prediction quality among the compared methods. On MMLU, the majority vote obtains the lowest NLL and Brier score, while Cooke weighting remains competitive. This pattern suggests that Cooke weighting is not uniformly dominant in homogeneous settings, but it offers a strong uncertainty-aware aggregation rule, especially on the more difficult MMLU-Pro benchmark. In contrast, accuracy weighting remains competitive in accuracy but is weaker on probabilistic metrics, reflecting its inability to distinguish confident and uncertain correct predictions.

Overconfident error further highlights the trade-offs between aggregation strategies. On MMLU, the majority vote is slightly higher at 2.14\%, while others all obtain the lowest OE. On MMLU-Pro, Cooke weighting achieves the lowest OE among the weighted methods at 2.34\%, close to global weighting at 2.35\% and below majority vote at 3.42\%. These results suggest that probability aggregation methods reduce overconfident errors relative to discrete voting through uniform averaging or estimated trust.

Overall, in the homogeneous setting, aggregation methods can appear competitive because all experts are strong, and suppression of confidence does not significantly harm accuracy.

\subsection{Heterogeneous expert panels}

Table~\ref{tab:heterogeneous results} presents results on MMLU and MMLU-Pro under the heterogeneous expert panel, where models exhibit substantial variation in capability and uncertainty quality. Compared to the homogeneous setting, this regime more clearly exposes differences between aggregation strategies.

\begin{table*}[htbp]
\centering
\small
\begin{tabular}{llcccccccc}
\toprule
& \multicolumn{4}{c}{MMLU} & \multicolumn{4}{c}{MMLU-Pro} \\
\cmidrule(lr){2-5} \cmidrule(lr){6-9}
Method  & Acc $\uparrow$ & NLL $\downarrow$ & Brier $\downarrow$ & OE $\downarrow$
              & Acc $\uparrow$ & NLL $\downarrow$ & Brier $\downarrow$ & OE $\downarrow$ \\
\midrule
Best solo model (Claude) & 93.50 & 0.329 & 0.143 & 3.27 & \textbf{87.97} & 0.601 & \textbf{0.231} & 3.29 \\
Majority vote & 92.27 & \textbf{0.303} & 0.143 & 3.20 & 84.22 & 0.645 & 0.259 & 4.97 \\
Equal averaging & 92.10 & 0.369 & 0.166 & 1.68 & 83.81 & 0.636 & 0.277 & \textbf{1.76} \\
Global weighting & 93.266 & 0.341 & 0.146 & 1.68 & 86.99 & 0.583 & 0.246 & 1.79 \\
Accuracy weighting & 92.72 & 0.358 & 0.157 & 1.75 & 85.30 & 0.613 & 0.264 & 1.82 \\
\textbf{Cooke weighting} & \textbf{94.27} & 0.331 & \textbf{0.139} & \textbf{1.65} & 87.78 & \textbf{0.558} & \textbf{0.231} & 1.82 \\
\bottomrule
\end{tabular}
\caption{Results of heterogeneous panels including 7 hybrid models on MMLU and MMLU-Pro. Values are averaged over 5 runs; standard deviations are negligible and omitted (see Table~\ref{tab:full result hete panel} in Appendix). Best results are in Bold.}
\label{tab:heterogeneous results}
\end{table*}

A similar trend is observed in both MMLU and MMLU-Pro, Cooke weighting achieves the strongest overall performance among ensemble methods, substantially outperforming majority vote and equal averaging. In particular, Cooke achieves the highest accuracy on MMLU, surpassing the best solo model and all other ensemble methods. These results indicate that uncertainty-aware trust estimation becomes beneficial once expert heterogeneity is introduced, in contrast to the homogeneous setting where gains were limited.

Cooke weighting consistently yields the best or near-best probabilistic metrics across both datasets. On MMLU-Pro, it achieves the lowest NLL (0.558) and Brier score (0.231).
On MMLU, it gets the best Brier score (0.139) and maintains a competitive NLL. In contrast, the majority vote achieves a relatively strong NLL on MMLU (0.303), but this does not translate to robustness under the more challenging MMLU-Pro setting, where its NLL degrades substantially (0.645). This suggests that methods that ignore uncertainty may appear competitive on easier datasets but fail under increased difficulty.

The results of overconfident error rates demonstrate that equal averaging achieves the lowest OE (1.76) on MMLU-Pro and Cooke weighting achieves similarly low OE (1.65 on MMLU, 1.82 on MMLU-Pro) while maintaining significantly higher accuracy.
The majority vote exhibits a higher OE (3.20 on MMLU), reflecting its inability to control overconfident failures.

\begin{figure}[htbp]
    \centering
    \includegraphics[width=\linewidth]{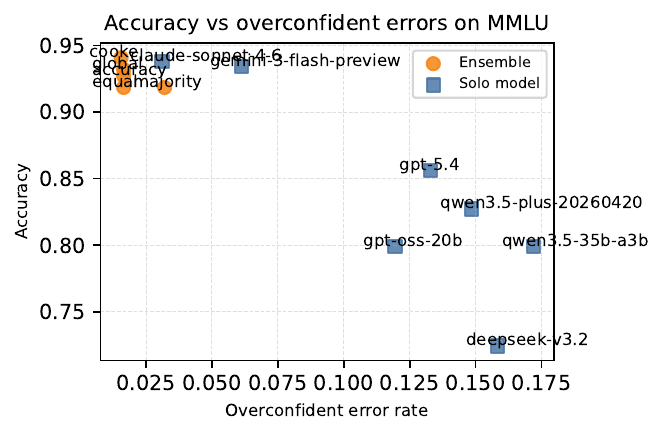} \vfill
  \includegraphics[width=\linewidth]{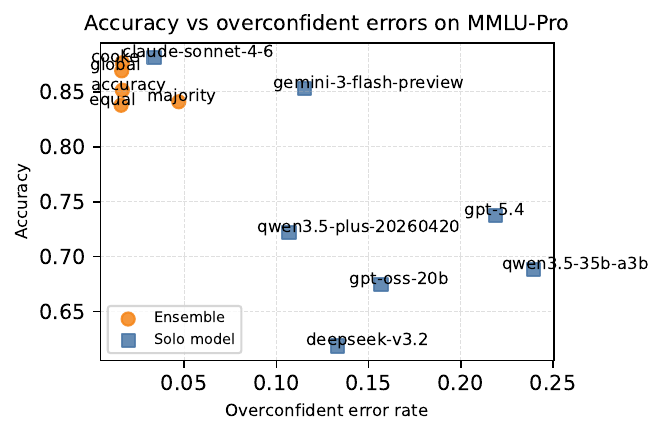}
  \caption {Accuracy vs overconfidence trade-off across aggregation methods and solo models. Methods in the upper-left region are preferable, as they achieve high accuracy while avoiding overconfidence errors.}
  \label{fig:accuracy_oe}
\end{figure}

To better understand the trade-off between predictive performance and uncertainty reliability, we plot accuracy against overconfident error rate in Figure~\ref{fig:accuracy_oe}. The results show that several methods achieve similar accuracy but differ greatly in overconfident errors. Strong solo models can be highly accurate, yet often produce more confident errors than ensemble methods. This indicates that aggregation can improve the risk profile of predictions even when accuracy gains are small.

Cooke weighting occupies a favourable region of the plot. It remains competitive in accuracy while maintaining a low overconfident error rate. This is expected because log-score weighting penalizes experts that assign low probability to correct seed answers, especially when they are confidently wrong. As a result, Cooke weighting provides a better balance between correctness and confident mistakes under heterogeneous expert panels.

\subsection{Contaminated panels}

We now evaluate robustness under contaminated expert panels, where a subset of experts is unreliable or adversarial. This setting is designed to test whether an aggregation method can distinguish reliable experts from unreliable ones. We systematically vary the contamination ratio by progressively adding noisy experts to real LLM panels, and measure how aggregation methods degrade as the proportion of unreliable experts increases. 

\begin{figure}[htbp]
  \includegraphics[width=\linewidth]{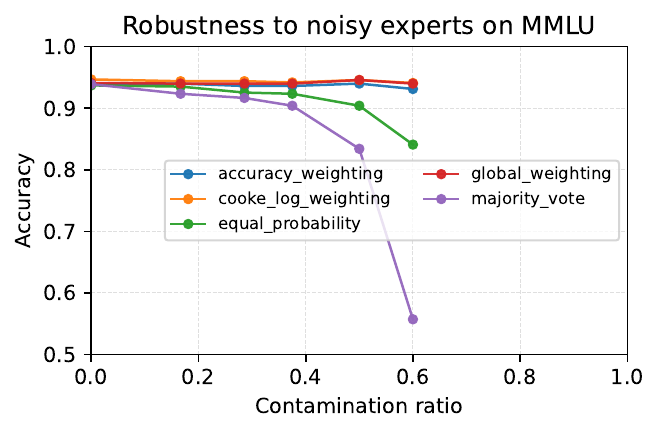} \vfill
  \includegraphics[width=\linewidth]{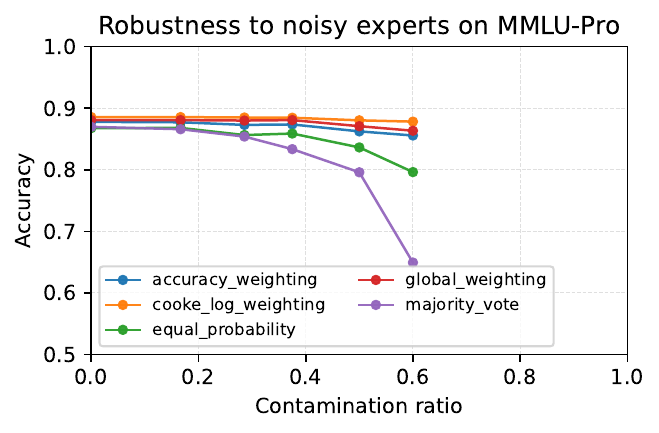}
  \caption{Robustness curve as contamination ratio increases.}
    \label{fig:robustness curve}
\end{figure}

Figure~\ref{fig:robustness curve} shows the accuracy as contamination increases on two datasets. In particular, majority voting is highly sensitive to contamination. As noisy experts accumulate, majority decisions are increasingly dominated by unreliable predictions, leading to substantial performance degradation. Equal averaging exhibits a steady decline in accuracy due to uniform smoothing, which suppresses both correct and incorrect signals indiscriminately. In contrast, calibration-based weighting (global and accuracy weighting) methods degrade more slowly, indicating that seed question performance provides useful information for suppressing unreliable experts. Cooke weighting maintains the strongest performance under high contamination ratios.

\begin{figure}
    \centering
    \includegraphics[width=\columnwidth]{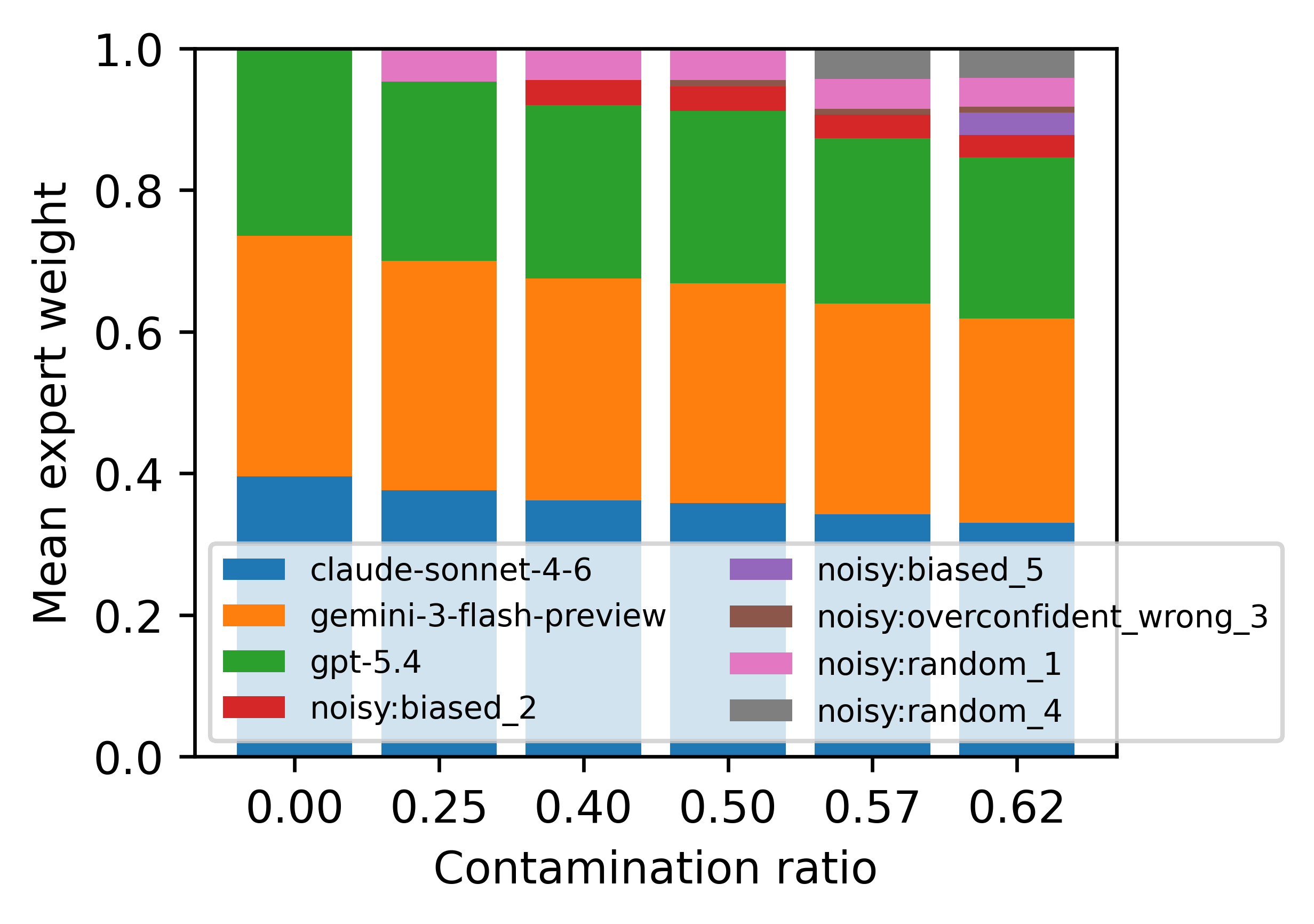}
    \caption{Weight distribution over different models as contamination ratio increases, tested on the MMLU-Pro.}
    \label{fig:weighting}
\end{figure}

To understand the mechanism behind the robustness curve, we visualise how Cooke weighting reallocates trust as the contamination ratio increases in Figure~\ref{fig:weighting}. Cooke weighting does not allocate weight to noisy experts in proportion to their count. Instead, most of the weights remain concentrated on the real LLM experts, while the noisy experts receive comparatively small weights. This shows it reallocates trust toward experts with stronger calibrated seed performance. 

Overall, the contaminated panel shows that calibration-based weighting improves robustness by limiting the influence of unreliable experts. 
However, the advantage of Cooke weighting is modest compared to the global weighting. This is because the synthetic noisy experts used there are globally unreliable: they perform poorly across nearly all subjects. 

Therefore, to test whether the Cooke weighting can localise trust when expertise is domain-specific, we introduce two subject-specific contamination types: 1) specialist: which is highly accurate on a designated subject but unreliable elsewhere, and 2) corrupted expert: which is generally reliable but confidently wrong on a designated subject. This setting directly tests the mechanism Cooke weighting is designed to capture: expert reliability that varies across domains.

We fix the real expert pool to three LLMs and inject five synthetic subject-specific experts, yielding a contamination ratio of \(5/(3+5)=62.5\%\). This creates a deliberately challenging setting in which synthetic experts form the majority of the ensemble. The specialist experts target biology, business, and economics, while the corrupted experts target economics.

Table~\ref{tab:contamination subject} shows the targeted subject results on MMLU-Pro. Cooke weighting performs best on all subjects in both the specialist and corrupted conditions, improving greatly over global weighting. Overall accuracy follows the same pattern (Table~\ref{tab:overall contamination subject}).

\begin{table}[htbp]
\centering
\small
\resizebox{\columnwidth}{!}{
\begin{tabular}{@{}lllllllll@{}}
\toprule
Setting & Target Subject & Majority & Equal & Global & Accuracy & \textbf{Cooke}   \\ \midrule
Specialist & Biology & 96.86 & 98.08 & 94.60 & 98.26 & \textbf{98.95}   \\
Specialist & Business & 95.25 & 97.94 & 88.43 & 99.37 & \textbf{100.00}   \\
Specialist & Economics & 91.70 & 93.48 & 90.22 & 94.52 & \textbf{96.44}  \\
Corrupted & Economics & 71.56 & 67.41 & 84.74 & 85.78 & \textbf{88.89}  \\ \bottomrule
\end{tabular}
}
\caption{Targeted subject accuracy on contaminated panel consisting of 3 real LLM experts and 5 subject-specific experts on MMLU-Pro dataset.}
\label{tab:contamination subject}
\end{table}

\begin{table}[htbp]
\resizebox{\columnwidth}{!}{
\small
\begin{tabular}{@{}lllllll@{}}
\toprule
Setting & Majority & Equal & Global & Accuracy & \textbf{Cooke}  \\ \midrule
Specialist & 89.74 & 92.62 & 89.39 & 93.32 & \textbf{94.17}  \\
Corrupted & 93.25 & 92.20 & 94.59 & 96.63 & \textbf{97.36}  \\ \bottomrule
\end{tabular}
}
\caption{Overall accuracy on contaminated panel consisting of 3 real LLM experts and 5 subject-specific experts on MMLU-Pro dataset.}
\label{tab:overall contamination subject}
\end{table}

\begin{figure}[htbp]
    \centering
    \includegraphics[width=\columnwidth]{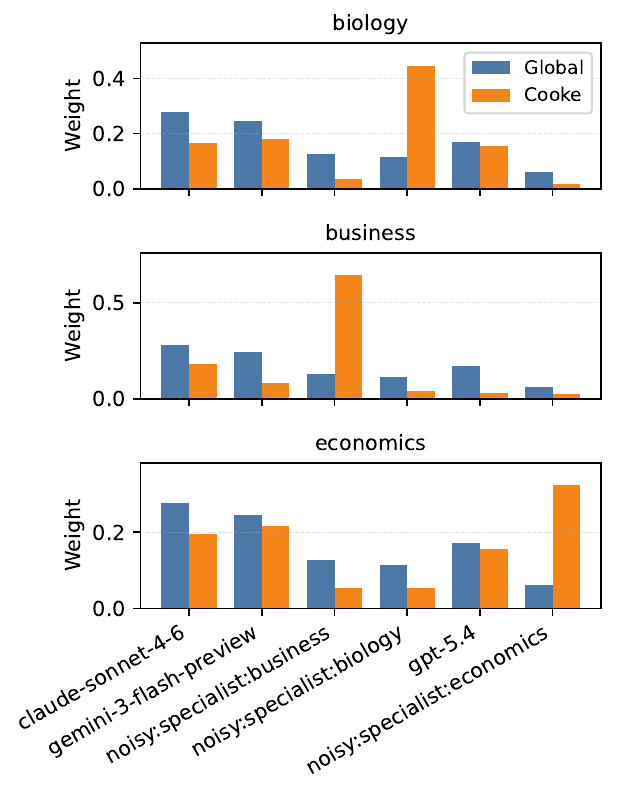}
    \caption{Weight distribution across experts by subject over Global weighting and Cooke weighting.}
    \label{fig:weight subject}
\end{figure}

Figure~\ref{fig:weight subject} visualises the corresponding weight allocation in the specialist condition. Global weights are constant across subjects, while Cooke weights vary by subject. Cooke concentrates weight on the relevant specialist experts for biology, business, and economics, while assigning them little weight outside their domains. In the corrupted setting, Cooke sharply downweights the corrupted economics experts on economics, but does not suppress them uniformly on unrelated subjects. These results show the main benefit of context-aware trust estimation introduced by Cooke weighting.

\subsection{Ablation analysis}

We finally analyse the sensitivity of calibration-based aggregation to the amount of seed data. Since Cooke weighting estimates expert reliability from seed questions, its effectiveness depends on whether the seed set is large enough to produce stable calibration estimates.

\begin{figure}[htbp]
    \centering
    \includegraphics[width=\columnwidth]{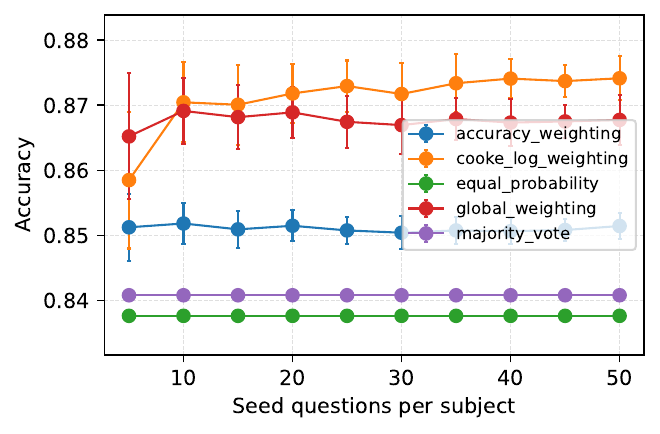}
    \caption{Seed question size ablation analysis on heterogeneous expert panels on MMLU-Pro.}
    \label{fig:seed size}
\end{figure}

Figure~\ref{fig:seed size} shows the seed-size ablation in the heterogeneous expert panel on the MMLU-Pro dataset. Cooke weighting benefits most from additional seeds, improving from 85.85\% accuracy with five seed questions per subject to 87.42\% with fifty. Global weighting is more stable, increasing only slightly from 86.52\% to 86.78\%, while accuracy weighting remains nearly flat around 85.1\%. These results suggest that small seed sets are already sufficient to estimate a useful model trust, but larger seed sets improve the trust estimation in Cooke weighting.

\section{Discussion}

Our results suggest that Cooke weighting provides the most balanced aggregation strategy, rather than uniformly dominating every metric. Majority vote and equal averaging are simple and competitive when all experts are strong, but they assume equal reliability and become fragile when noisy experts enter the pool. Similarly, the disagreement-based aggregation method~\cite{kruse-etal-2025-simple} assumes that disagreement reflects uncertainty. While it can effectively capture ensemble uncertainty, disagreement does not necessarily correspond to reliability, especially in contaminated settings, where adversarial experts can either align with each other or coincide with incorrect consensus.

Accuracy weighting can be more conservative, producing slightly lower performance accuracy and probabilistic quality. Compared with accuracy weighting, Cooke is less conservative but more informative: it can distinguish between models that make the same accuracy but assign very different probabilities to the correct answer. Global weighting is a strong baseline because it estimates model reliability from seed questions, and it performs well when expert quality is relatively stable across subjects.

The advantage of Cooke weighting becomes clearer under contamination. When noisy experts are globally unreliable, both global and Cooke weighting can identify them from seed questions. However, when expertise or failure is subject-specific, Cooke weighting provides a stronger mechanism: it can upweight specialists only where they are reliable and downweight corrupted experts only where they fail. This context-aware trust estimation is the main empirical benefit of structured expert judgment in LLM ensembles.

The ablation results further suggest that Cooke weighting does not require a large seed set to be useful, but additional seed questions improve the stability of subject-level estimates. 
Overall, the evidence supports a measured claim: Cooke-style structured expert judgment achieves the best overall accuracy-calibration-robustness balance, providing a mechanism for domain-specific trust estimation.

\section{Conclusion}

We formulate multi-LLM aggregation as uncertainty-aware trust estimation, showing that existing accuracy-centric aggregation methods are insufficient under heterogeneity and contamination. By leveraging structured expert judgment and proper scoring rules, our approach improves robustness and achieves a superior accuracy-reliability balance. These results highlight the value of calibrating trust under uncertainty in multi-LLM systems.

\section*{Limitations}

Our approach relies on calibration (seed) questions to estimate expert reliability. While we show consistent improvements across settings, performance may depend on how well the seed set reflects the target distribution. If the seed set does not reflect the target distribution, the estimated weights may not generalise well. Our method also assumes access to probabilistic outputs from each model, e.g., obtained via verbalised confidence. Their quality can vary across models, and inaccuracies in probability estimates may affect weighting. We focus on multiple-choice benchmarks, where probability distributions are naturally defined. Extending the approach to open-ended generation tasks remains an important direction for future work. Finally, our contamination experiments simulate unreliable experts in a controlled manner. While this setting enables systematic evaluation, real-world failure modes may be more complex. Further study is needed to assess robustness in more diverse deployment scenarios.

\section*{Information about AI use}
We used AI tools (e.g., ChatGPT) to assist with editing and improving the clarity of the writing. All technical content, experimental design, and analysis were developed and verified by the authors.



\bibliography{custom}

\appendix

\section{Appendix}
\label{sec:appendix}

\subsection{Evaluation metrics}

Accuracy: we report standard top-1 probability accuracy, formally:
\begin{equation}
    \text{Acc}
=
\frac{1}{|T|}
\sum_{i \in T}
\mathbf{1}[\hat{y}_i = y_i^\ast]
\end{equation}
where $\hat{y}_i$ and $y_i^\ast$ represent the predicted choice with the highest probability and the ground truth, respectively.

Negative Log Likelihood (NLL): NLL measures how much probability the ensemble assigns to the correct answer, formally:
\begin{equation}
    \text{NLL}
=
-\frac{1}{|T|}
\sum_{i \in T}
\log p_{\text{agg}}(y_i^\ast \mid x_i)
\end{equation}
Lower NLL indicates that a method assigns a higher probability to the correct answer.

Brier score: Brier score measures the squared error between predicted probabilities and the one-hot correct label. 
Lower Brier values indicate better probabilistic predictions. Compared with NLL, the Brier score is less sensitive to extremely small probabilities assigned to the true answer. Formally: 
\begin{equation}
\small
   \text{Brier}
=
\frac{1}{|T|}
\sum_{i \in T}
\sum_{y \in \mathcal{Y}}
\left(
p_{\text{agg}}(y_i \mid x_i)
-
\mathbf{1}[y_i = y_i^\ast]
\right)^2
\end{equation}

\subsection{Experimental setup details}

The system prompt for probability elicitation is:

\begin{lstlisting}[basicstyle=\small\ttfamily,breaklines=true,columns=fullflexible]
You are answering a multiple-choice benchmark question. 
Return calibrated probabilities over all answer choices.
Return only a JSON object with exactly these keys:{A,B,C,D}
The values must be numbers summing to 1. 
\end{lstlisting}


The prompt asks the model to output a JSON object whose keys are the valid answer labels, e.g., A through D for MMLU or A through J for MMLU-Pro. 

The prompt template for answering each question is:


\begin{lstlisting}[basicstyle=\small\ttfamily,breaklines=true,columns=fullflexible]
Question: {question}
Respond as JSON only, for example: {A: 0.25, B: 0.25, C: 0.25, D: 0.25}
Return exactly this format and nothing else.
\end{lstlisting}


\section{More results}

\subsection{Homogeneous frontier panels}

\begin{table*}[t]
\centering
\resizebox{\linewidth}{!}{
\scriptsize
\begin{tabular}{lcccccccccc}
\toprule
& \multicolumn{5}{c}{MMLU} & \multicolumn{5}{c}{MMLU-Pro} \\
\cmidrule(lr){2-6} \cmidrule(lr){7-11}
Method & Acc. & NLL & Brier & ECE & OE & Acc. & NLL & Brier & ECE & OE \\
\midrule
Cooke weighting
& 94.04$_{\pm 0.20}$ & 0.265$_{\pm 0.005}$ & 0.109$_{\pm 0.004}$ & 0.097$_{\pm 0.002}$ & 1.92$_{\pm 0.13}$
& \textbf{88.03$_{\pm 0.43}$} & \textbf{0.485$_{\pm 0.006}$} & \textbf{0.201$_{\pm 0.003}$} & 0.114$_{\pm 0.004}$ & \textbf{2.34$_{\pm 0.13}$} \\

Global weighting
& 94.17$_{\pm 0.21}$ & 0.271$_{\pm 0.005}$ & 0.113$_{\pm 0.003}$ & 0.100$_{\pm 0.002}$ & 1.92$_{\pm 0.13}$
& 86.86$_{\pm 0.74}$ & 0.502$_{\pm 0.006}$ & 0.212$_{\pm 0.004}$ & 0.113$_{\pm 0.008}$ & 2.35$_{\pm 0.09}$ \\

Accuracy weighting
& 94.14$_{\pm 0.16}$ & 0.273$_{\pm 0.005}$ & 0.116$_{\pm 0.003}$ & 0.099$_{\pm 0.001}$ & 1.92$_{\pm 0.13}$
& 86.45$_{\pm 0.36}$ & 0.505$_{\pm 0.007}$ & 0.217$_{\pm 0.004}$ & 0.106$_{\pm 0.001}$ & 2.43$_{\pm 0.09}$ \\

Majority vote
& \textbf{94.19$_{\pm 0.31}$} & \textbf{0.249$_{\pm 0.006}$} & \textbf{0.107$_{\pm 0.003}$} & \textbf{0.044$_{\pm 0.002}$} & 2.14$_{\pm 0.14}$
& 86.99$_{\pm 0.26}$ & 0.583$_{\pm 0.017}$ & 0.216$_{\pm 0.005}$ & \textbf{0.051$_{\pm 0.002}$} & 3.42$_{\pm 0.18}$ \\

Equal probability
& 93.86$_{\pm 0.26}$ & 0.277$_{\pm 0.005}$ & 0.119$_{\pm 0.003}$ & 0.099$_{\pm 0.001}$ & 1.92$_{\pm 0.13}$
& 86.02$_{\pm 0.31}$ & 0.516$_{\pm 0.007}$ & 0.225$_{\pm 0.004}$ & 0.111$_{\pm 0.001}$ & 2.42$_{\pm 0.10}$ \\

\midrule
Claude Sonnet 4.6
& 93.50$_{\pm 0.18}$ & 0.330$_{\pm 0.004}$ & 0.134$_{\pm 0.002}$ & 0.108$_{\pm 0.002}$ & 3.32$_{\pm 0.19}$
& 87.77$_{\pm 0.33}$ & 0.605$_{\pm 0.006}$ & 0.233$_{\pm 0.003}$ & 0.143$_{\pm 0.002}$ & 3.34$_{\pm 0.09}$ \\

Gemini 3 Flash
& 93.75$_{\pm 0.33}$ & 0.304$_{\pm 0.007}$ & 0.119$_{\pm 0.005}$ & \textbf{0.026$_{\pm 0.002}$} & 5.77$_{\pm 0.32}$
& 85.21$_{\pm 0.49}$ & 0.778$_{\pm 0.022}$ & 0.268$_{\pm 0.008}$ & 0.080$_{\pm 0.002}$ & 11.47$_{\pm 0.35}$ \\

GPT-5.4
& 85.30$_{\pm 0.61}$ & 0.581$_{\pm 0.019}$ & 0.262$_{\pm 0.010}$ & 0.067$_{\pm 0.008}$ & 13.55$_{\pm 0.57}$
& 73.55$_{\pm 0.36}$ & 1.151$_{\pm 0.010}$ & 0.454$_{\pm 0.005}$ & 0.137$_{\pm 0.003}$ & 21.72$_{\pm 0.26}$ \\
\bottomrule
\end{tabular}
}
\caption{Full results of homogeneous panel on MMLU and MMLU-Pro. Values report mean$_{\pm}$standard deviation over 5 trials. Accuracy and overconfident error rate (OE) are reported as percentages. Lower is better for NLL, Brier, ECE, and OE.}
\label{tab:apx:top3models}
\end{table*}

Tables~\ref{tab:apx:top3models} report the full main results for the top-3 frontier model ensembles on MMLU and MMLU-Pro, including all solo models and ensemble methods. In addition to accuracy, NLL, Brier score, and overconfident error rate, we include expected calibration error (ECE) to measure how closely predicted confidence matches empirical correctness. On MMLU, Cooke weighting ties global weighting for the best accuracy while achieving the lowest NLL and Brier score, indicating stronger probabilistic quality. On MMLU-Pro, Cooke weighting achieves the best ensemble accuracy and the lowest NLL and Brier score, although the strongest solo model remains slightly higher in raw accuracy. These results show that Cooke weighting is especially competitive when evaluation considers both correctness and calibrated probability estimates, rather than accuracy alone.

\subsection{Heterogeneous panels}

\begin{table*}[hbtp]
\centering
\resizebox{\linewidth}{!}{
\scriptsize
\begin{tabular}{lcccccccccc}
\toprule
& \multicolumn{5}{c}{MMLU} & \multicolumn{5}{c}{MMLU-Pro} \\
\cmidrule(lr){2-6} \cmidrule(lr){7-11}
Method & Acc. & NLL & Brier & ECE & OE & Acc. & NLL & Brier & ECE & OE \\
\midrule
Cooke weighting
& \textbf{94.27$_{\pm 0.17}$} & 0.331$_{\pm 0.002}$ & 0.139$_{\pm 0.000}$ & 0.155$_{\pm 0.002}$ & \textbf{1.65$_{\pm 0.29}$}
& \textbf{87.78$_{\pm 0.30}$} & \textbf{0.558$_{\pm 0.006}$} & \textbf{0.232$_{\pm 0.003}$} & 0.173$_{\pm 0.002}$ & 1.82$_{\pm 0.09}$ \\

Global weighting
& 93.66$_{\pm 0.28}$ & 0.341$_{\pm 0.002}$ & 0.146$_{\pm 0.001}$ & 0.154$_{\pm 0.002}$ & 1.68$_{\pm 0.24}$
& 86.99$_{\pm 0.21}$ & 0.583$_{\pm 0.005}$ & 0.246$_{\pm 0.003}$ & 0.178$_{\pm 0.001}$ & 1.79$_{\pm 0.13}$ \\

Accuracy weighting
& 92.72$_{\pm 0.19}$ & 0.358$_{\pm 0.002}$ & 0.157$_{\pm 0.000}$ & 0.154$_{\pm 0.003}$ & 1.75$_{\pm 0.19}$
& 85.30$_{\pm 0.28}$ & 0.613$_{\pm 0.004}$ & 0.264$_{\pm 0.002}$ & 0.170$_{\pm 0.001}$ & 1.82$_{\pm 0.09}$ \\

Majority vote
& 92.27$_{\pm 0.30}$ & \textbf{0.303$_{\pm 0.009}$} & 0.143$_{\pm 0.001}$ & \textbf{0.073$_{\pm 0.005}$} & 3.20$_{\pm 0.10}$
& 84.22$_{\pm 0.19}$ & 0.645$_{\pm 0.013}$ & 0.259$_{\pm 0.003}$ & 0.074$_{\pm 0.002}$ & 4.97$_{\pm 0.08}$ \\

Equal probability
& 92.10$_{\pm 0.24}$ & 0.369$_{\pm 0.001}$ & 0.165$_{\pm 0.001}$ & 0.151$_{\pm 0.002}$ & 1.68$_{\pm 0.24}$
& 83.81$_{\pm 0.44}$ & 0.636$_{\pm 0.004}$ & 0.277$_{\pm 0.002}$ & 0.163$_{\pm 0.004}$ & \textbf{1.76$_{\pm 0.07}$} \\
\midrule
Claude Sonnet 4.6
& 93.50$_{\pm 0.17}$ & 0.329$_{\pm 0.003}$ & 0.143$_{\pm 0.002}$ & 0.108$_{\pm 0.002}$ & 3.27$_{\pm 0.15}$
& \textbf{87.97$_{\pm 0.19}$} & 0.601$_{\pm 0.005}$ & 0.231$_{\pm 0.003}$ & 0.144$_{\pm 0.001}$ & 3.29$_{\pm 0.07}$ \\

Gemini 3 Flash
& 93.62$_{\pm 0.31}$ & 0.307$_{\pm 0.006}$ & \textbf{0.122$_{\pm 0.004}$} & \textbf{0.026$_{\pm 0.001}$} & 5.92$_{\pm 0.17}$
& 85.52$_{\pm 0.25}$ & 0.763$_{\pm 0.010}$ & 0.263$_{\pm 0.004}$ & 0.079$_{\pm 0.001}$ & 11.23$_{\pm 0.17}$ \\

GPT-5.4
& 85.02$_{\pm 0.31}$ & 0.590$_{\pm 0.009}$ & 0.268$_{\pm 0.005}$ & 0.070$_{\pm 0.005}$ & 13.85$_{\pm 0.30}$
& 73.78$_{\pm 0.18}$ & 1.144$_{\pm 0.001}$ & 0.451$_{\pm 0.001}$ & 0.135$_{\pm 0.002}$ & 21.57$_{\pm 0.06}$ \\

DeepSeek V3.2
& 73.56$_{\pm 0.44}$ & 0.956$_{\pm 0.015}$ & 0.428$_{\pm 0.008}$ & 0.072$_{\pm 0.003}$ & 14.34$_{\pm 0.39}$
& 61.86$_{\pm 0.25}$ & 1.589$_{\pm 0.008}$ & 0.586$_{\pm 0.003}$ & 0.141$_{\pm 0.001}$ & 12.88$_{\pm 0.18}$ \\

GPT-OSS 20B
& 79.13$_{\pm 0.26}$ & 0.717$_{\pm 0.008}$ & 0.316$_{\pm 0.003}$ & 0.052$_{\pm 0.000}$ & 13.20$_{\pm 0.35}$
& 67.19$_{\pm 0.55}$ & 1.404$_{\pm 0.028}$ & 0.500$_{\pm 0.010}$ & 0.128$_{\pm 0.006}$ & 16.69$_{\pm 0.47}$ \\

Qwen3.5 35B
& 80.29$_{\pm 0.35}$ & 0.919$_{\pm 0.015}$ & 0.369$_{\pm 0.005}$ & 0.128$_{\pm 0.002}$ & 16.99$_{\pm 0.26}$
& 69.17$_{\pm 0.17}$ & 1.757$_{\pm 0.023}$ & 0.559$_{\pm 0.005}$ & 0.208$_{\pm 0.006}$ & 23.58$_{\pm 0.44}$ \\

Qwen3.5 Plus
& 83.56$_{\pm 0.37}$ & 0.648$_{\pm 0.012}$ & 0.292$_{\pm 0.006}$ & 0.038$_{\pm 0.003}$ & 14.14$_{\pm 0.31}$
& 72.31$_{\pm 0.28}$ & 1.110$_{\pm 0.007}$ & 0.435$_{\pm 0.004}$ & \textbf{0.050$_{\pm 0.004}$} & 10.53$_{\pm 0.04}$ \\
\bottomrule
\end{tabular}
}
\caption{Full results of heterogeneous panel on MMLU and MMLU-Pro. Values report mean$_{\pm}$standard deviation over 5 trials. Accuracy and overconfident error rate (OE) are reported as percentages. Lower is better for NLL, Brier, ECE, and OE.}
\label{tab:full result hete panel}
\end{table*}

Table~\ref{tab:full result hete panel} shows the full result on heterogeneous panels.

\begin{figure}[htbp]
    \centering
    \includegraphics[width=\linewidth]{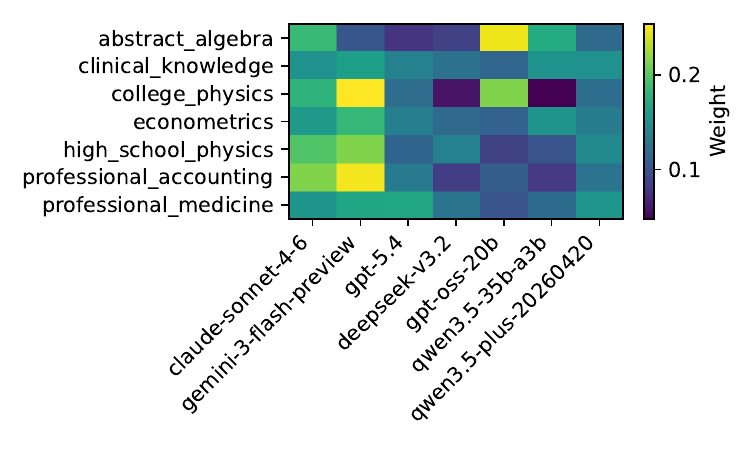}
    \caption{Weight distribution in heterogeneous panel on MMLU.}
    \label{fig:weight distribution MMLU.}
\end{figure}

\begin{figure}[htbp]
    \centering
    \includegraphics[width=\linewidth]{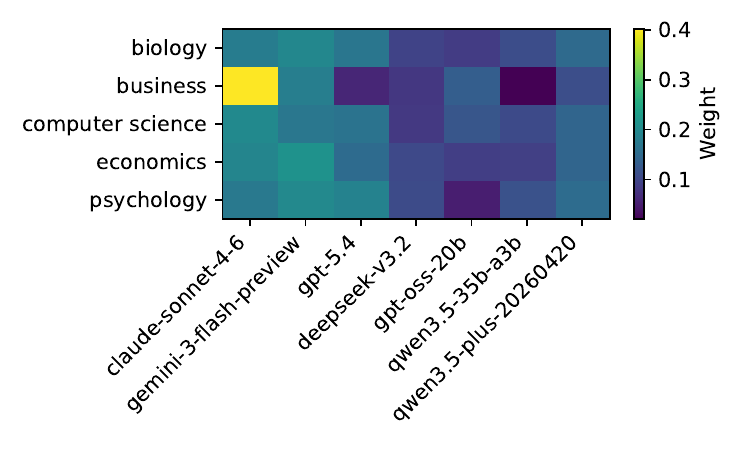}
    \caption{Weight distribution in heterogeneous panel on MMLU-Pro.}
    \label{fig:weight distribution MMLU-Pro.}
\end{figure}

Figure~\ref{fig:weight distribution MMLU.} and ~\ref{fig:weight distribution MMLU-Pro.} visualise the weight distribution across subjects on the heterogeneous panel. The results show that Cooke weighting assigns most of the weight to the strongest real LLMs rather than spreading the weight uniformly across all seven models. On MMLU, the weights are concentrated on the top-performing models, i.e., Claude and Gemini. The same pattern appears on MMLU-Pro. Because the benchmark is harder, and model performance is more uneven, Cooke weighting places more trust in the top 3 frontier models.

\begin{figure*}[htbp]
    \centering
    \begin{subfigure}{0.49\linewidth}
        \includegraphics[width=\linewidth]{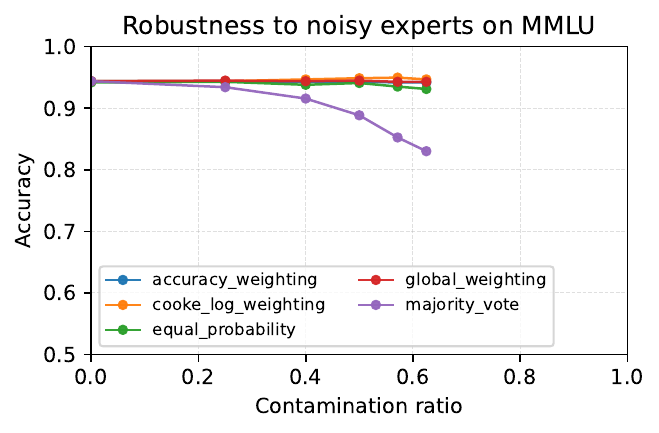}
        \caption{Random experts.}
    \end{subfigure}
    \hfill
    \begin{subfigure}{0.49\linewidth}
        \includegraphics[width=\linewidth]{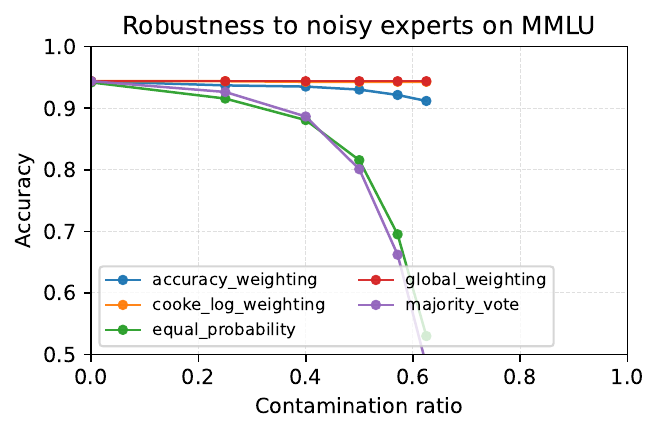}
        \caption{Overconfident wrong experts.}
    \end{subfigure}
    \vspace{-0.5em}
    \begin{subfigure}{0.49\linewidth}
        \includegraphics[width=\linewidth]{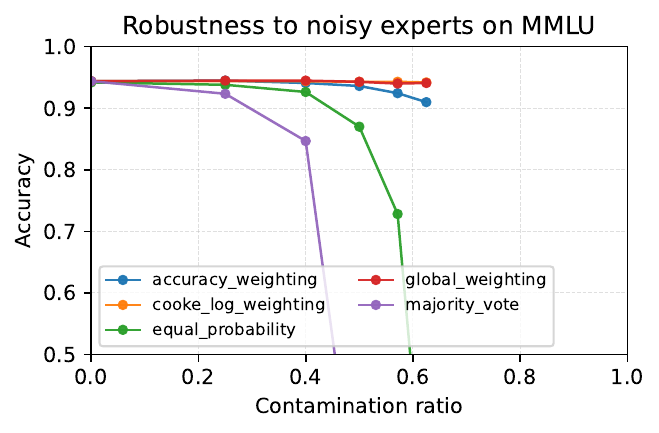}
        \caption{Biased experts}
    \end{subfigure}
    \hfill
    \begin{subfigure}{0.49\linewidth}
        \includegraphics[width=\linewidth]{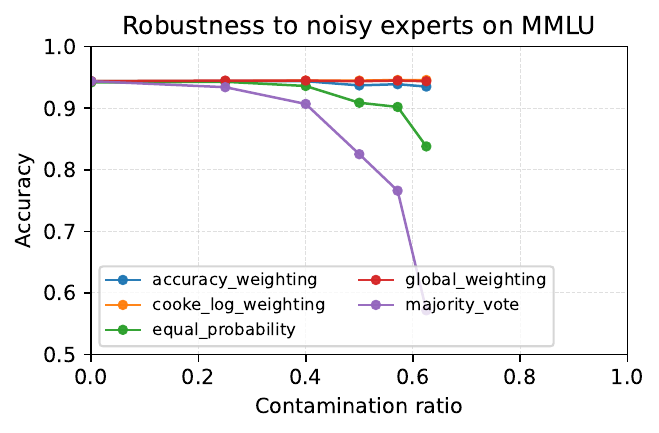}
        \caption{Mixed of three types.}
    \end{subfigure}
    \caption{Robustness curves under different noisy expert types on MMLU.}
    \label{fig:apd:robustness}
\end{figure*}

\subsection{Contaminated panels}

Figure~\ref{fig:apd:robustness} visualises the robustness curves under different noisy expert types on MMLU. As the contamination ratio increases, unweighted aggregation methods generally degrade more rapidly. In contrast, seed-based weighting methods are more stable, indicating that calibration examples provide useful information for identifying unreliable experts before target aggregation. The degradation is most pronounced for biased and overconfident-wrong experts, where majority vote and equal averaging are particularly vulnerable. Cooke weighting remains competitive across all contamination settings, showing that probabilistic seed scoring can suppress poorly calibrated experts.

\begin{figure}[tbp]
    \centering
    \includegraphics[width=\linewidth]{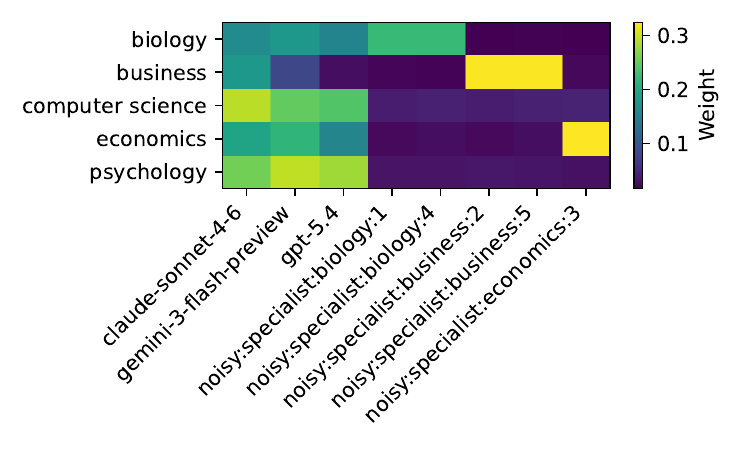}
    \caption{Weight distribution on contaminated panels with specialist experts.}
    \label{fig:weight spec experts}
\end{figure}

\begin{figure}[tbp]
    \centering
    \includegraphics[width=\linewidth]{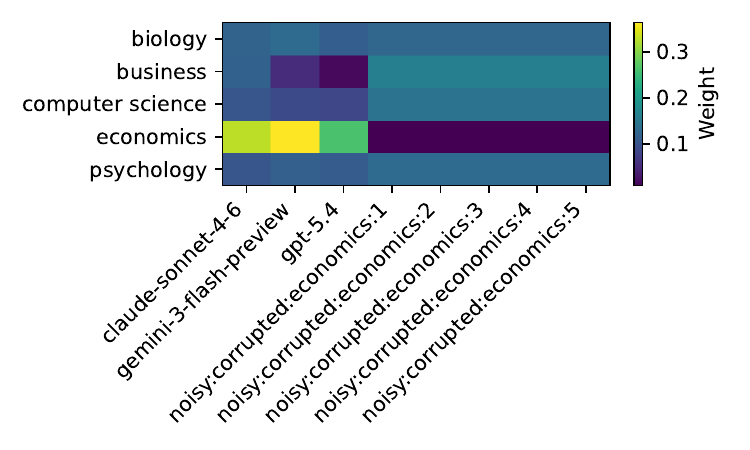}
    \caption{Weight distribution on contaminated panels with a corrupted economics expert.}
    \label{fig:weight corr experts}
\end{figure}

Figure~\ref{fig:weight spec experts} and~\ref{fig:weight corr experts} visualise the weight distribution on subject-level contaminated panels with specialist and corrupted experts. The results show that Cooke-style weighting adapts its trust allocation by subject. In the specialist setting, the method assigns high weight to specialist experts on their corresponding target subjects, while giving them much lower weight on unrelated subjects. In the corrupted setting, the opposite pattern emerges. The corrupted experts receive non-trivial weight on subjects where they remain reliable, but are sharply downweighted on the corrupted target subject.

\subsection{Ablation analysis}

\begin{figure}[htbp]
    \centering
    \includegraphics[width=\linewidth]{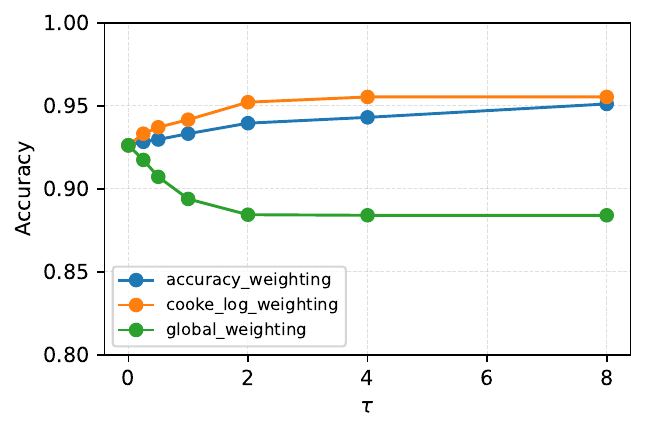}
    \caption{Ablation analysis of $\tau$ on accuracy in subject-level contaminated panels.}
    \label{fig:accuracy tau}
\end{figure}

\begin{figure}[htbp]
    \centering
    \includegraphics[width=\linewidth]{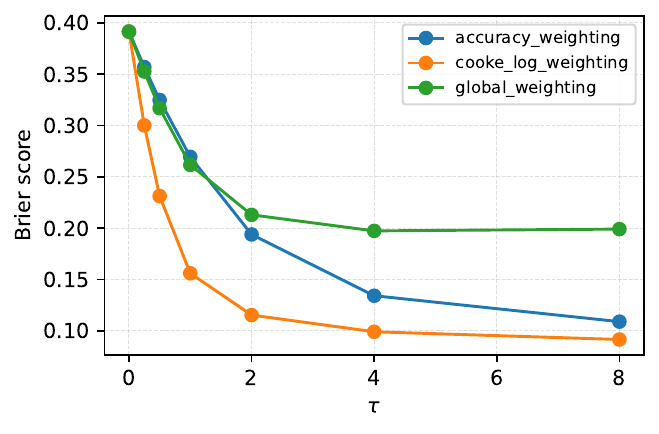}
    \caption{Ablation analysis of $\tau$ on Brier score in subject-level contaminated panels.}
    \label{fig:brier tau}
\end{figure}

\begin{figure}[htbp]
    \centering
    \includegraphics[width=\linewidth]{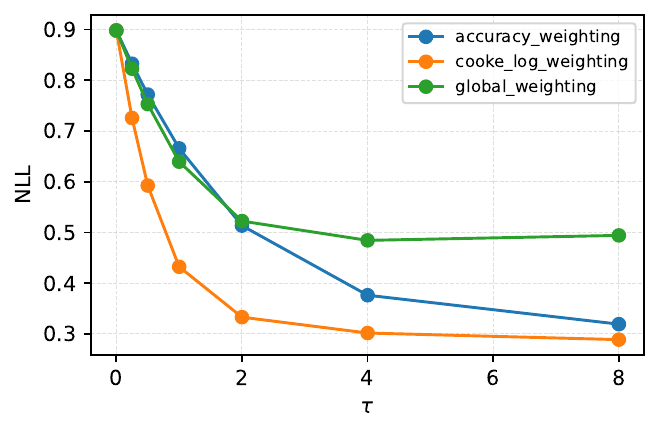}
    \caption{Ablation analysis of $\tau$ on NLL in subject-level contaminated panels.}
    \label{fig:NLL tau}
\end{figure}

We analyse how the value of $\tau$ affects the final accuracy and other probabilistic quality metrics on the subject-level contaminated panels. It examines how sharply seed scores are converted into ensemble weights. The results are shown in Figure~\ref{fig:accuracy tau}, ~\ref{fig:brier tau}, and~\ref{fig:NLL tau}. When \(\tau=0\), all softmax-based weighting methods reduce to uniform averaging, so Cooke, global, accuracy weighting, and equal averaging coincide. The results show that moderate values of \(\tau\) improve performance by allowing the ensemble to favour better-calibrated models, while very large values can over-concentrate weight and reduce robustness.

\end{document}